\documentclass{article}
\usepackage[final,nonatbib]{neurips_2024}
\usepackage[utf8]{inputenc} % allow utf-8 input

\usepackage[T1]{fontenc}    % use 8-bit T1 fonts
\usepackage{hyperref}       % hyperlinks
\usepackage{url}            % simple URL typesetting
\usepackage{booktabs}       % professional-quality tables
\usepackage{amsfonts}       % blackboard math symbols
\usepackage{nicefrac}       % compact symbols for 1/2, etc.
\usepackage{microtype}      % microtypography
\usepackage{lipsum}         % Can be removed after putting your text content
\usepackage{graphicx}
\usepackage{doi}

\usepackage{amsmath}
\usepackage{graphicx}
\usepackage{url}            % simple URL typesetting
\usepackage{booktabs}       % professional-quality tables
\usepackage{nicefrac}       % compact symbols for 1/2, etc.
\usepackage{microtype}      % microtypography
\usepackage{xcolor}         % colors

\usepackage{latexsym}
\usepackage{inconsolata}
\usepackage{multirow}
\usepackage{bbding}

\usepackage{comment}
\usepackage{amsmath,amssymb,amsfonts}

\usepackage[ruled]{algorithm}
\usepackage{algpseudocode}
\usepackage{caption}
\usepackage{xfrac}

\usepackage{comment}

\begin{document}

\title{Collaboratively adding new knowledge to an LLM}

\author{
Rhui Dih Lee \\
IBM Research \\
\texttt{ rhui.dih.lee@ibm.com }
\\ \And 
Laura Wynter \\
IBM Research \\
\texttt{ lwynter@sg.ibm.com }
}

\maketitle
%%%%%%%%%%%%%====== Abstract  ==============
\begin{abstract}
 We address the question of how to successively add new knowledge to an LLM whilst retaining previously-added knowledge. We consider  two settings, semi-cooperative and fully-cooperative. Overall, LoRA  performs better in most cases than full-fine tuning of all parameters when both new knowledge acquisition and old, including recent, knowledge retention are taken into account. In the semi-cooperative setting, where datasets are not  available after training, MOE mixing, model merging, and LoRA-based orthogonal subspace sequential learning, using a small weight on the orthogonality term, perform well.  In the fully-cooperative setting where datasets remain available, joint training and sequential training with replay are both effective approaches with LoRA training  generally preferable to full fine-tuning.

\end{abstract}

%%%%%%%%%%%%%======   ==============
\section{Introduction}
With the availability of open weight large language models (LLM) such as the  popular llama models \cite{llama}, along with  easy-to-use libraries for fine-tuning such as Hugging Face Transformers \cite{HF, HFurl}, training an LLM on one's own data has become a standard practice. Typically, each use case results in a different fine-tuning of the source model, resulting in many tuned models obtained from the same base model. 

LLM tuning aims to give new knowledge or skills to the LLM. The LLM should not, however, lose its ability to perform well on the tasks it was previously trained for, including tasks pertaining to  the general knowledge from pre-training and skills like instruction following.
This tug-of-war is further amplified when the LLM is successively subjected to more tuning, over time. 
While one can return to the base model for each successive tuning, users would like to be able to leverage on recent knowledge from previous tuning, and further tune from there.

The options available for tuning the base model are essentially limited to full fine-tuning (FFT) the model or using Parameter-Efficient Fine Tuning (PEFT). In the former case, the user is left with multiple copies of a potentially large LLM, each tailored to a different use case. In the latter case, the user can maintain one copy of the base model and a set of smaller adapters, e.g. LoRA adapters, one for each use case. The choice   as to whether to use FFT or PEFT in this setting then resolves around the quality of the tuned models obtained versus the computational resources needed to store the resulting models.

If, however, the user wishes to reap benefits from  recently-acquired knowledge, in addition to the  new knowledge that the user will impart in training the model, then further options become available. A terminology used in the pre-LLM era for this is "continual learning". In continual learning, a model is sequentially subjected to training on new data,  over time,  without reverting back to the original base model.

The risk inherent in subjecting a base model to sequential training on different domains is that the old knowledge from pre-training the model, as well as the recently-acquired knowledge from earlier training, will be degraded. There is thus an intrinsic trade-off, or tug-of-war, when fine-tuning a model. The question we wish to answer is how to best win this tug-of-war between old knowledge degradation and new knowledge acquisition.

An important  distinction in this context of  sequential training, is whether users have access to the earlier datasets when performing training on their own dataset. When the training is fully collaborative, all datasets used in training are centralised and remain accessible. In semi-collaborative settings, the models resulting from successive training remain accessible, but not the datasets used to train them.

We address both settings, the fully-collaborative and the semi-collaborative setting, and provide an answer to the question of how to best enable successive accumulation of knowledge by an LLM while minimising the degradation of old and recently-acquired knowledge.

The paper proceeds as follows. First we cover the necessary background and related work. Then, in Section \ref{sec:methods} we define the methods that we explore. Section \ref{sec:exp} presents the experimental setup and  the results. Finally, we conclude with a discussion in \ref{sec:conc}.

% \href{https://www.overleaf.com/learn}{help library}

\section{Related Work}
\label{sec:relatedwork}

Full fine-tuning a large language model (LLM) involves updating potentially all of its parameters. The seminal work  of \cite{lora} introduced LoRA, i.e.,  Low-Rank Adaptation, a form of parameter-efficient fine-tuning (PEFT), which demonstrated that remarkably good results can be achieved while training only a very small number of parameters associated with the linear layers of the LLM. The success of LoRA paved the way for  more efficient training of LLMs.

In parallel, however, the development and widespread use of LoRA also led to questions about whether fine-tuning such a small number of parameters from the base model, instead of all the parameters, would help to reduce the degradation of old knowledge that can arise from  training. In \cite{loralearnsless}, the authors discuss that question in detail. They conclude that 
LoRA results in less knowledge acquisition than full fine-tuning (FFT), but that LoRA better retains the
base model performance from old knowledge. 

The authors of \cite{unsloth-lora}, however, claim that the LoRA results in the paper \cite{loralearnsless} were not as good as they could be due to the choice of hyper-parameters  used. Specifically, they  claim that the implementation of LoRA used in \cite{loralearnsless}, which only trained the attention, and up and down matrices, and  not  the gate projection matrix, performed poorly because the latter is important in obtaining good results. Secondly, they claim that the  LoRA training should have included the lm\_head and embed\_tokens  to take into account domain data distribution shift. They state, however, that one must use a smaller learning rate for the lm\_head and embed\_tokens, and show that training those layers is very sensitive and can result in worse performance in some cases. They also point out that the LoRA rank used in the code tests was $r=256$ with an $\alpha$ of 32, whereas for the math tests, $r=256$ and  $\alpha=2r$. They claim that the higher $\alpha$ in the presence of large rank explains the better math results and should also have been used in the code tests. 
The results of \cite{loralearnsless,unsloth-lora} are important in assessing the benefits and disadvantages of full fine-tuning versus LoRA PEFT. However, we are concerned with how to successively add new knowledge to a model while benefiting from the knowledge added already by earlier fine-tuning. In this context either FFT or PEFT can be used. 

The authors of  \cite{trace} analyse model performance across numerous models looking primarily at sequential FFT, sequential LoRA training, and sequential FFT with replay. However, the results provided in the paper are mixed and somewhat conflicting.  They provide a suite of datasets for training and testing that we make use of in this work.

Finally, we mention a related work on LoRA training for successive tasks which seeks to drive the adapter to be orthogonal to previously-trained adapters, called o-lora \cite{olora}. We examine whether this technique is of value in the setting where a user does not have access to datasets used in previous training. 

\section{Methods}
\label{sec:methods} 

We consider both  full fine tuning, or FFT, and  LoRA tuning. For sequential FFT,  the previous training checkpoint must be available. For sequential LoRA, the previous adapters, along with the full base model, must be available.

In the semi-collaborative setting, datasets from previous training are not available. Sequential training, whether FFT or LoRA, can be performed. Orthogonal-subspace o-lora \cite{olora} is  a feasible method as well, whereby successive LoRA adapters are encouraged to be orthogonal to preceding adapters by modifying the loss function to be:
\begin{equation}
\sum_{x,y\in D_t} \log p_{\Theta}(y|x) + \lambda \sum{i=1}^{t-1} L_{orth}(A_i,A_t),
\end{equation}
where 
\begin{equation}
L_{orth}(A_i,A_t)= \sum_{j,k} \|O_{i,t}[j,k]\|^2.
\end{equation}
The orthogonality term thus aims to enforce $O_{i,t}=A_i^T A_t=0$ to an extent depending upon the parameter $\lambda$,  where $O_{i,t}[j,k]$ denotes the element at the $j$th row
and $k$th column of $O_{i,t}$. This method aims to reduce interference between successive LoRA training, postulated to be useful when the tasks of each training are quite different from each other. Note that the orthogonality term $L_{orth}$ in the loss function acts as a soft constraint, and modifies the optimization of the CE loss. As such, we  expect that the learning ability of the model trained using o-lora will be reduced in order to achieve the reduced interference with previously-trained adapters. 

In addition, two other options are available which do not require access to the datasets used in training the previous models: namely, model merging and MOE model mixing. Model merging refers to averaging the parameters of the model, and a set of methods that extends simple averaging to handle various cases including parameters with opposite signs. The reader may refer to the survey paper \cite{merging} on model merging for more details. In this work, we consider simple parameter averaging as well as the merging method TIES and a combination of TIES \cite{ties} and DARE \cite{dare}.

MOE model mixing is yet a different method to combine already-trained models without requiring access to the source data used for training them. Model mixing takes one trained model as the base, and several other trained models as experts. The experts contribute the FFN layers (and potentially also the attention layers) in a parallel manner and  routers are added to route tokens to a subset of the expert modules. The architecture is the same as that of Mixtral \cite{mixtral} without requiring training of the expert modules or the base.
The authors of \cite{moe} propose several variants for MOE model mixing, some of which require datasets for training the routers, and others which do not. Training the routers, if desired, is the same as any continued pre-training or supervised fine-tuning, where only the parameters of the routers are able to be modified. Note that  datasets used for training the routers need not be the same as those used in training the models. The authors also enable the same approach to be performed using adapters as experts instead of full-parameter models, and release an   \href{https://github.ibm.com/Rhui-Dih-Lee/moetify}{open-source repo}.

 In the fully-collaborative setting, all of the datasets used previously for training the model remain available. In this setting,  we consider  joint training as an alternative to sequential training. Finally for the fully-collaborative setting, we consider sequential FFT and sequential LoRA training with replay in addition to standard sequential training. Replay is defined as adding a percentage, in our case, 10\%, of all previous task data into the current training. Note then that the dataset grows with each successive training. Note also that orthogonal subspace learning, o-lora, is  no longer appropriate since previous tasks' dataset are used in joint and sequential training with replay.

\section{Experimental Results}
\label{sec:exp} 

\subsection{Experimental Setup}

We make use of the TRACE \cite{trace} data as part of our experimental setup. The TRACE repository includes 8 tasks, each of which has a 5K-example dataset for training and a 2K-example dataset for testing. The  tasks include ScienceQA, which comprises elementary and high school science question-answer pairs, 
 FOMC,  a financial hawkish-dovish classification task, MeetingBank,   a
 city council meeting summarization task, and Py150 a python code-completion task.
It also includes two multi-lingual tasks, namely cstance,   a Chinese-language
 zero-shot stance detection task from Chinese social media postings,  and 20Minuten, a German news  text simplification task. 
 
 The TRACE dataset also includes two math tasks that we do not use. Instead we    make use  of MetaMath \cite{metamath}, a much larger 395K-example dataset, for math problem solving extended pre-training, and the two math solving-ability tests,  GSM8K and GSM8K\_COT \cite{gsm8k}. Hence we consider in total 7 tasks, 6 from TRACE and a math task using a much larger dataset than those in TRACE.

We also make use of evaluation tests from the LM Eval Harness \cite{eval-harness}, which includes tests such as MMLU, arc-challenge, commonsenseQA, hellaswag, openbookQA, piQA, and winogrande, that reflect the ability of the model to perform well on old knowledge, including the knowledge from pre-training. We call these tests collectively "Old Knowledge". Experiments are performed on one node with 8 A100-80GB GPU. We use llama3 (\texttt{meta-llama/Meta-Llama-3-8B}) \cite{llama3} for all tests reported here.

\subsection{One-Round Model Training}

\begin{figure}
\centering
\includegraphics[width=0.85\linewidth]{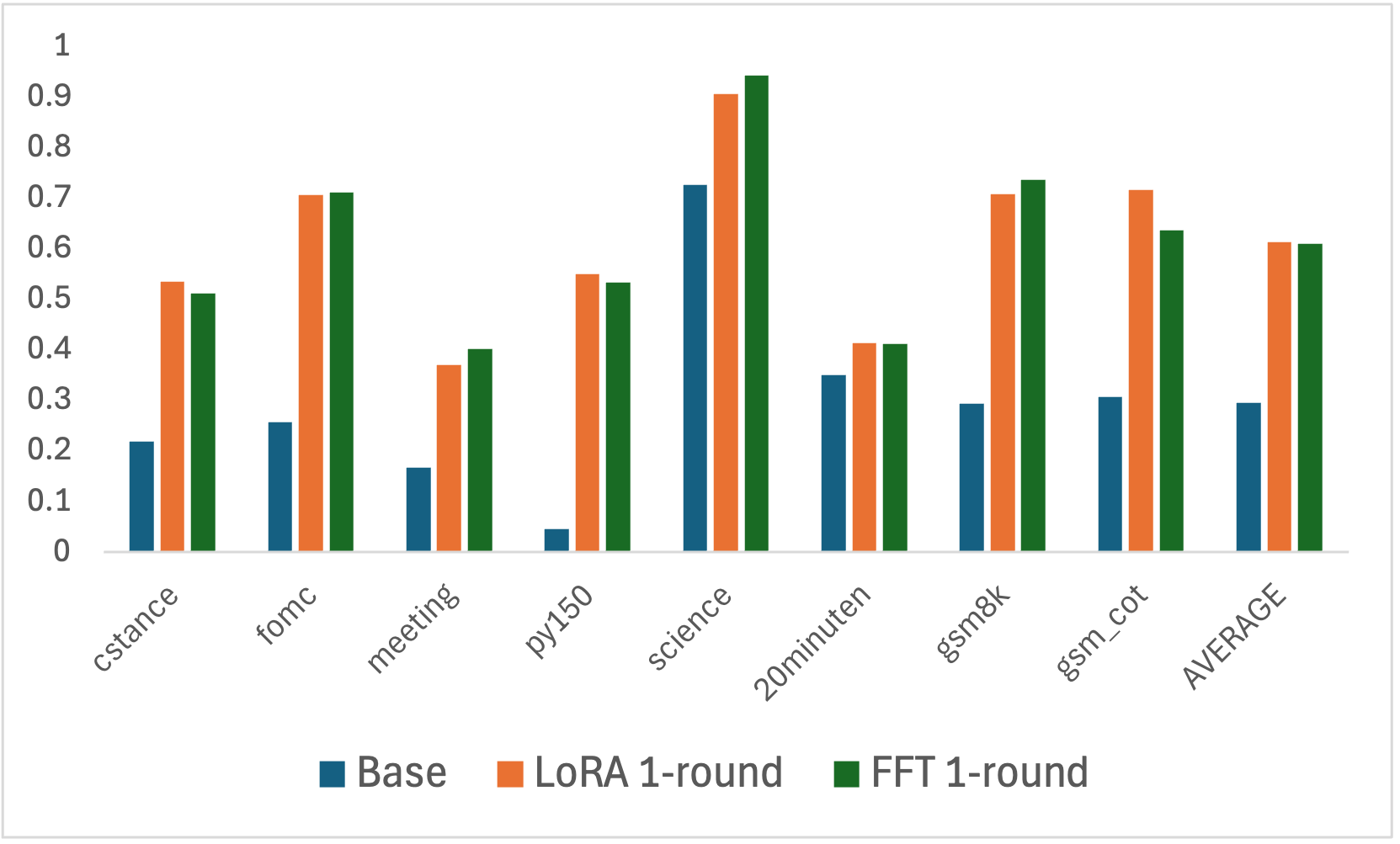}
\caption{One-round training, comparing  the base model, called "base" in the figure, with performance after fine-tuning the model on 6 different dataset-tasks. New knowledge acquisition on average is  similar with LoRA and FFT.}
\label{fig:1roundNew}
\end{figure}

We first consider individual, or one-round, training. In this case, all training starts from the base model. 
Figures \ref{fig:1roundNew}  illustrates the quality of the output after  one-round training  on the new knowledge the model is trained with. Observe that science question-answer and meeting summarization perform better when the model is fully fine-tuned. On the other hand, on cstance, the Chinese task, and python code completion the LoRA training performed better. The differences, however, in new knowledge acquisition are not very large between FFT and LoRA training, as shown in the last set of bars on the right which provide the average performance. 

\begin{figure}
\centering
\includegraphics[width=0.85\linewidth]{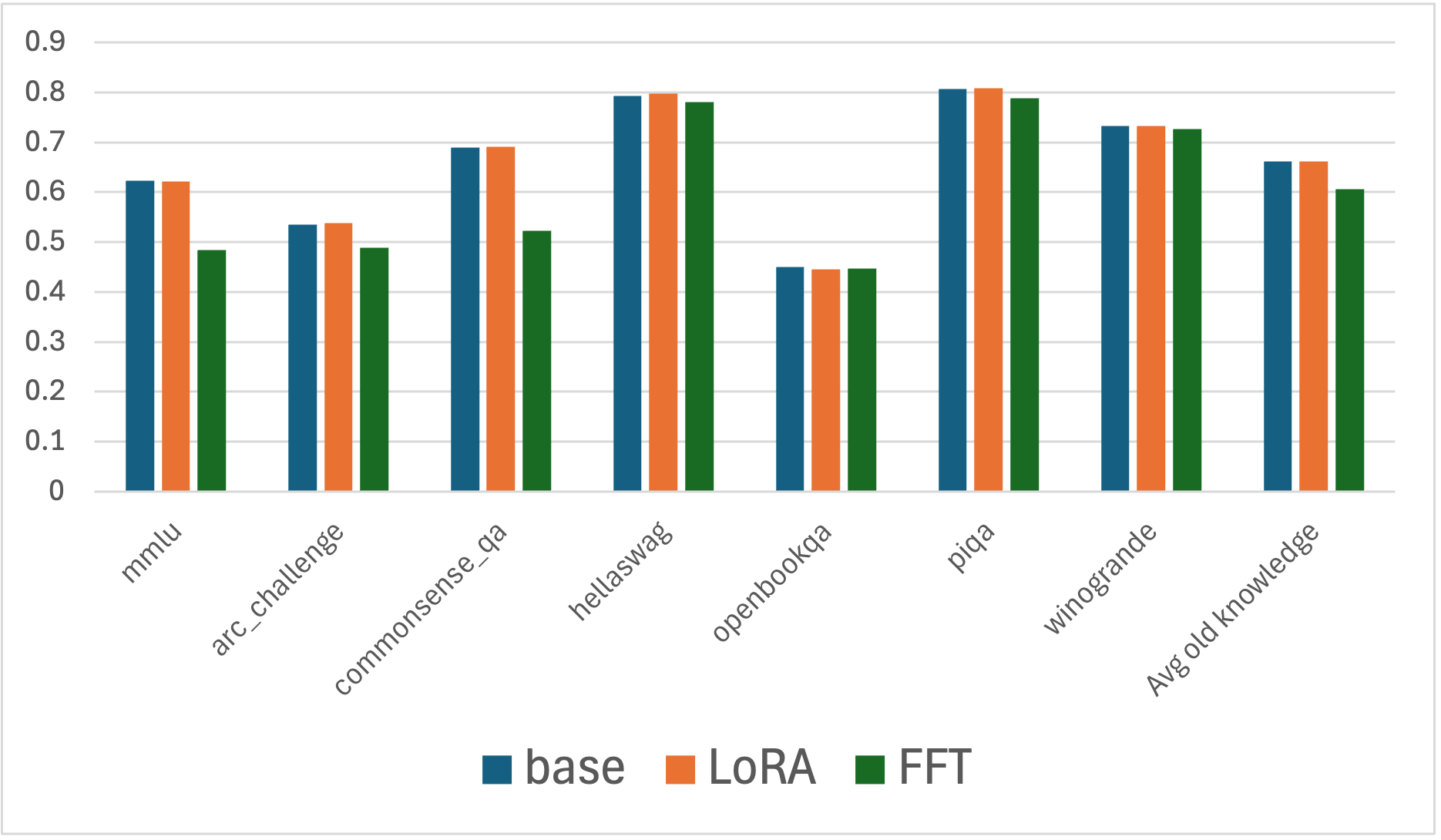}
\caption{Average over the one-round trained models, which were trained from the base model, and each tuned  on one  dataset-tasks. Average old-knowledge retention after one-round training across models, using full fine-tuning (FFT) suffers considerably more degradation then with LoRA training. Performance of the original base model is shown in the figure as "base".}
\label{fig:1roundOld}
\end{figure}

On the other hand, as shown in Figure  \ref{fig:1roundOld}, old knowledge retention suffers considerably when the training modifies potentially all parameters using full fine-tuning. We take each of the one-round-trained models and assess  old knowledge retention through the LM eval harness \cite{eval-harness} suite. Then, we average the results on a per-test basis over the one-round-trained models.
We see old knowledge degradation notably  on the tasks MMLU and CommonsenseQA which degrade more after FFT  than after LoRA training. The final set of bars provides an average over the tests of the model-average scores.

%%%%%%%%%%%%%%%
\subsection{Semi-Collaborative Model Training}

Turning next to semi-collaborative, sequential training. In this case, the datasets of previous training are not available to users, but the trained models and LoRA adapters remain available. 
As such, the user can start with the checkpoint from the previously-trained model and train from that tuned model instead of the base model.

It is clear that the order of training will have an impact on the resulting performance at any point in the chain of successive training. However, in practice, one may not be able to choose which checkpoint to start from and even if one could choose, the number of choices would become unwieldy as the number of successive training increases. For this reason, we do not explore the impact of task-order on the quality of the output. Instead, we maintain the same order as the authors of the TRACE dataset \cite{trace} and keep the order constant. The order of training is thus cstance, fomc, meeting, py150, science, 20minuten, and finally the very large math training dataset MetaMath.

\begin{figure}[ht]
\centering
\includegraphics[width=0.85\linewidth]{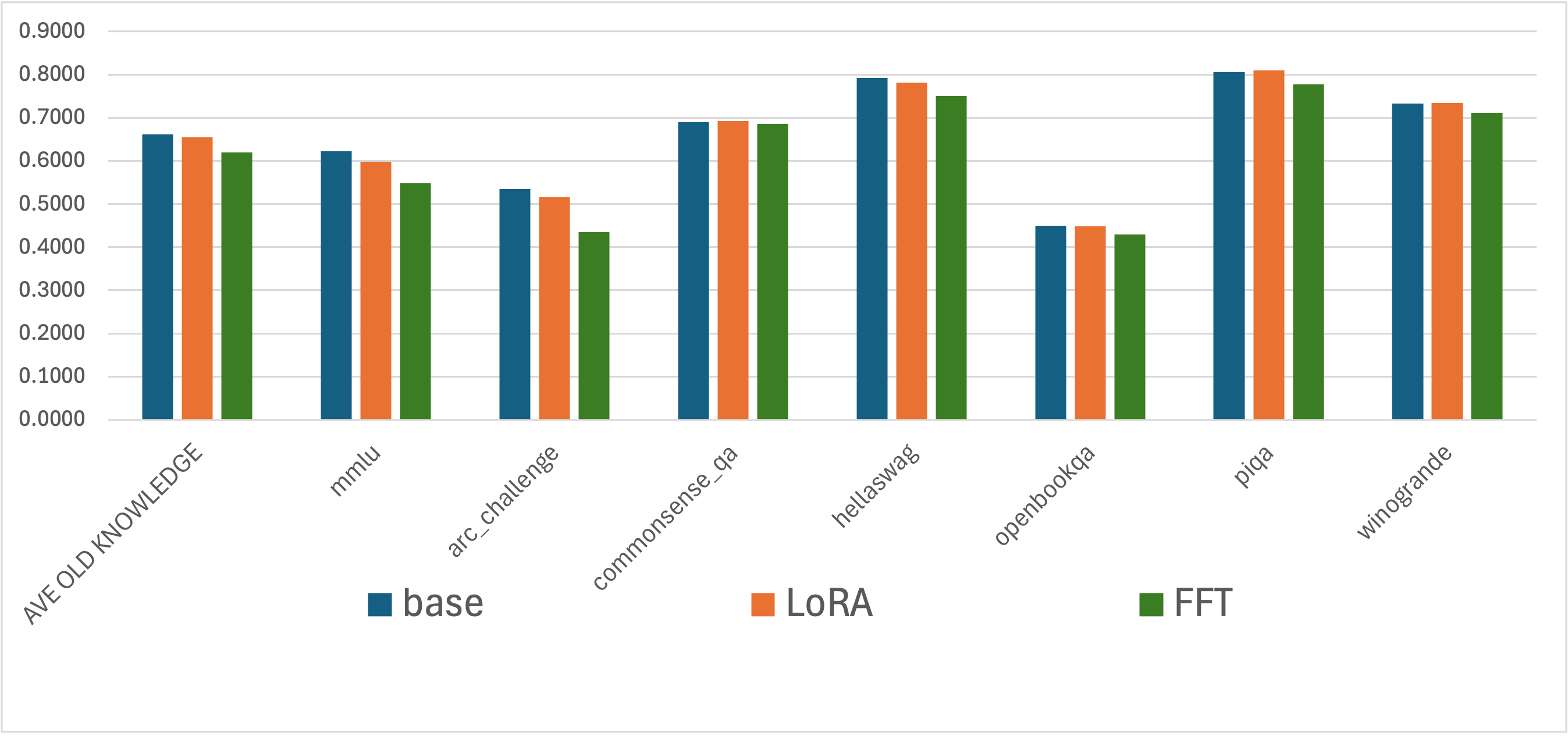}
\caption{Sequential training: old-knowledge retention after 7 rounds of sequential training, each training starting from the previously-trained model. Full fine-tuning suffers considerably more degradation of old knowledge than LoRA training. "Base" refers to the scores obtained by the original llama3.1 model without further training.}
\label{fig:7roundOld}
\end{figure}

We first examine the quality of old knowledge retention with successive training over the 7 datasets and tasks and compare against the scores obtained by the base model, that is, the original llama3.1 model without further training. Figure \ref{fig:7roundOld} summarises the results after all 7 rounds of training are complete. As we saw in one-round training, FFT, after successive training, clearly experiences  more degradation of old knowledge than LoRA training. 

\begin{figure}[ht]
\centering
\includegraphics[width=0.85\linewidth]{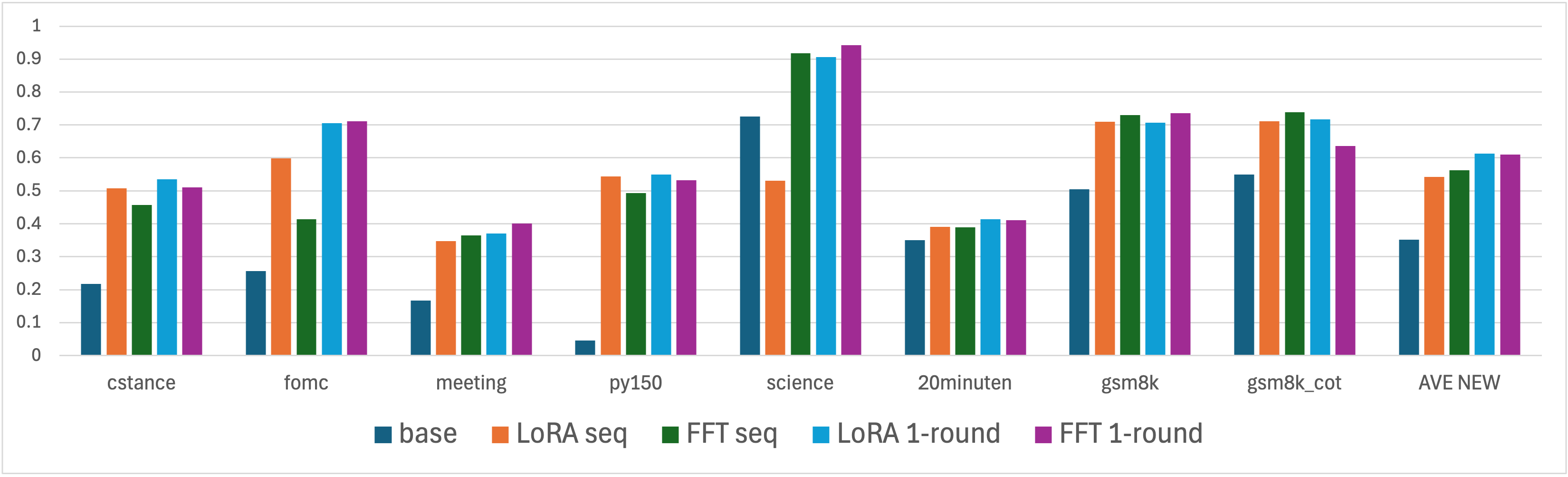}
\caption{Sequential training: new knowledge acquisition and retention after 7 rounds of  training, each training starting from the previously-trained model. Two right-most bars are one-round LoRA and one-round FFT training, which can be viewed as upper bounds on training performance. Sequential LoRA training is mostly adequate and comparable on average to sequential FFT on new data.}
\label{fig:7roundNew}
\end{figure}

In terms of new knowledge acquisition and recent knowledge retention, we examine the performance on all 7 tasks after the 7 rounds of sequential training.  Looking at Figure \ref{fig:7roundNew},   we see that  performance on average for recent knowledge acquisition and retention is comparable for LoRA and FFT, as shown in the last set of bars. The 4th and 5th bars in each set represent  one-round training on each dataset alone, hence without interference from subsequent tasks. As such the one-round results can be seen as an upper bound, to the extent that there is little synergy across the tasks in the TRACE datasets. Sequential training remains largely competitive with one-round training, and has the advantage of providing a single model than than an ever-larger growing number of separate models or adapters.

Sharp differences, however, can be seen on specific task performance, such as science, where sequential FFT performs dramatically better than LoRA, and Chinese, fomc classification and python code completion, where the reverse is true and LoRA provides better results. Overall, however, LoRA appears to be a better choice for sequential training than full-fine tuning for its better average knowledge acquisition and old knowledge retention.

Detailed performance on old knowledge is provided in Table \ref{tab:FFT} for full fine-tuning and Table \ref{tab:LoRA} for LoRA fine tuning, and for new knowledge acquisition and retention in Table \ref{tab:newFFT} for FFT and Table \ref{tab:newLoRA} for LoRA. Observe that overall the degradation from LoRA training is lower, at -1\% as compared to -7\% with FFT. BWT refers to backward transfer, defined as the percentage change in performance directly after training on the task vs.  the rounds of training after the task training. Note that as BWT shows the change due to subsequent training after a task's own training, BWT is not defined for the last task, which is math, as no training is performed after math.

In the semi-cooperative setting other paradigms exist in addition to standard sequential FFT and LoRA training.
Notice in Tables \ref{tab:newFFT} and \ref{tab:newLoRA} that successive training impacts very negatively the performance on some tasks, such as science, Specifically, the LoRA training on the math dataset caused a sharp decline in science performance of the final model. Similarly, training on subsequent tasks causes a sharp decline in performance on the fomc classification task. 

One way to mitigate this apparent conflict is to add an orthogonality term to the loss function as proposed in \cite{olora}. We consider orthogonal LoRA adapters  with a moderate weight of $\lambda=0.5$ on the orthogonality term and a version with a stronger weight of $\lambda=5$ on the orthogonality term. 

\begin{figure}[ht]
\centering
\includegraphics[width=0.85\linewidth]{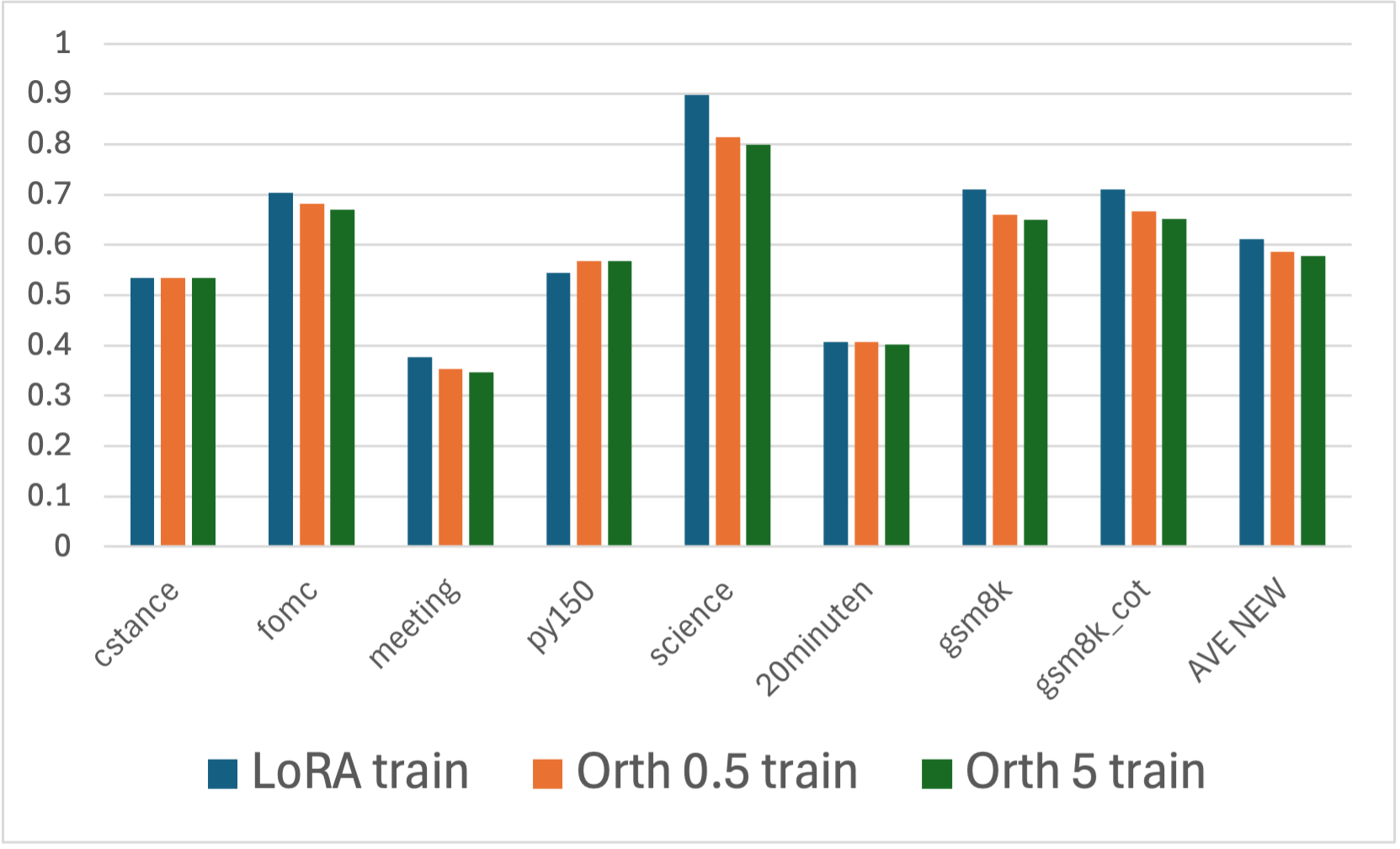}
\caption{Knowledge acquisition  quality after training using LoRA versus orthogonal LoRA with low ($\lambda=0.5$) and high ($\lambda=5$)weights. Quality is evaluated here immediately after training on the task in question before further training, hence only acquisition is represented here, not knowledge retention. }
\label{fig:olora-acq}
\end{figure}

As noted earlier, the addition of the orthogonality term in the loss function acts as a soft constraint, pushing the optimization away from pure CE loss minimization, so as to minimize conflicts with previously trained adapters. However, as a result, we can expect that o-lora training on a given task will not achieve as optimal a solution (in terms of CE loss) as a fully unconstrained optimization would achieve. Indeed, we see this in Figure \ref{fig:olora-acq} on most of the datasets. The figure shows the performance on each task immediately after training on that task, in a sequential manner. Note that this is not the final performance on all tasks at the end of the 7 rounds, but the performance on each task after its own training (Hence, these are the diagonal entries in the tables of sequential results). The  figure  highlights the new knowledge acquisition ability of LoRA versus o-lora.  On most tasks, the unconstrained LoRA achieves higher task scores, as is expected given that LoRA optimization is unconstrained by orthogonality requirements.

\begin{figure}[ht]
\centering
\includegraphics[width=0.85\linewidth]{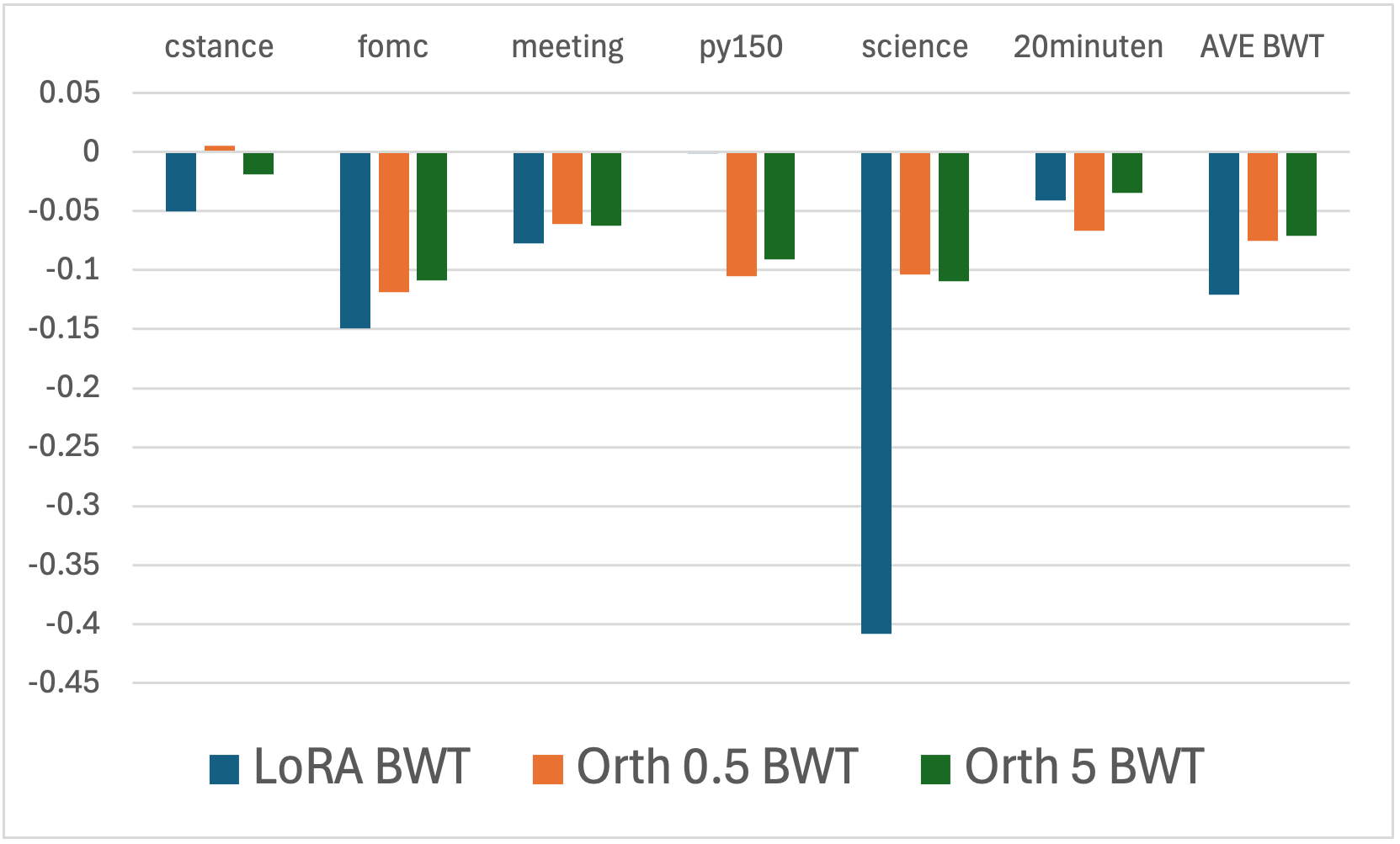}
\caption{Knowledge acquisition  and retention using LoRA versus orthogonal LoRA with low ($\lambda=0.5$) and high ($\lambda=5$) weights. Quality is evaluated here  after all 7 rounds of training, thus measuring recent knowledge retention. Note that earlier tasks in the sequence have greater chance for degradation than later tasks in the sequence, yet science near the end of the sequence suffers most.}
\label{fig:olora-bwt}
\end{figure}

On the other hand, the orthogonality of the o-lora-trained parameters indeed exhibit less interference with the previous adapter parameters, as can be seen in Figure \ref{fig:olora-bwt}. The figure shows the backward transfer (BWT) of each type of training at the end of the 7 rounds of training. The BWT is generally negative, as model performance on a past task tends to decrease when the model is subsequently trained on new tasks. Most striking is how large the negative BWT is on the science task using LoRA, and given that the science training is disrupted by only two other tasks after it (20minuten and math), the sharp degradation using LoRA is significant. The earlier tasks, cstance, fomc, and meeting also see significantly more loss in quality due to parameter interference. The choice as to using orthogonal o-lora versus LoRA thus depends on the likelihood of interference across subsequent tasks. 

It should be noted that the weight $\lambda$ on the orthogonality term can be adjusted for each sequential training, though we keep it constant in this study. As such the use of o-lora is a viable approach on a case-by-case basis whereby the weight can be made higher or lower at each successive training.

The net conclusion on the comparison between standard and orthogonal LoRA is that a small orthogonality weight can be advantageous in successive learning but should be considered on a case-by-case basis as it does limit the quality of the task-specific training. A good middle ground can be to use an orthogonality term with a small weight, $\lambda<=0.5$.

In addition to orthogonal subspace learning, two other approaches are available in the semi-cooperative setting: model merging and MOE model mixing. Both require only access to the trained models.While MOE model mixing can leverage datasets to train the routers, they need not be the datasets used in training the models or adapters. We consider both MOE Mixing of FFT models and MOE Mixing of LoRA adapters.

We also consider four types of model merging here, simple averaging of full model weights, called FFT merge in the figures, analogous averaging of LoRA adapter weights, called LoRA merge, the so-called TIES \cite{ties} model merging method, and the combination of TIES with another merging method, DARE \cite{dare}, both of which heuristically try to improve the averaging process by dropping some weights or accounting for different signs. 

\begin{figure}[ht]
\centering
\includegraphics[width=0.85\linewidth]{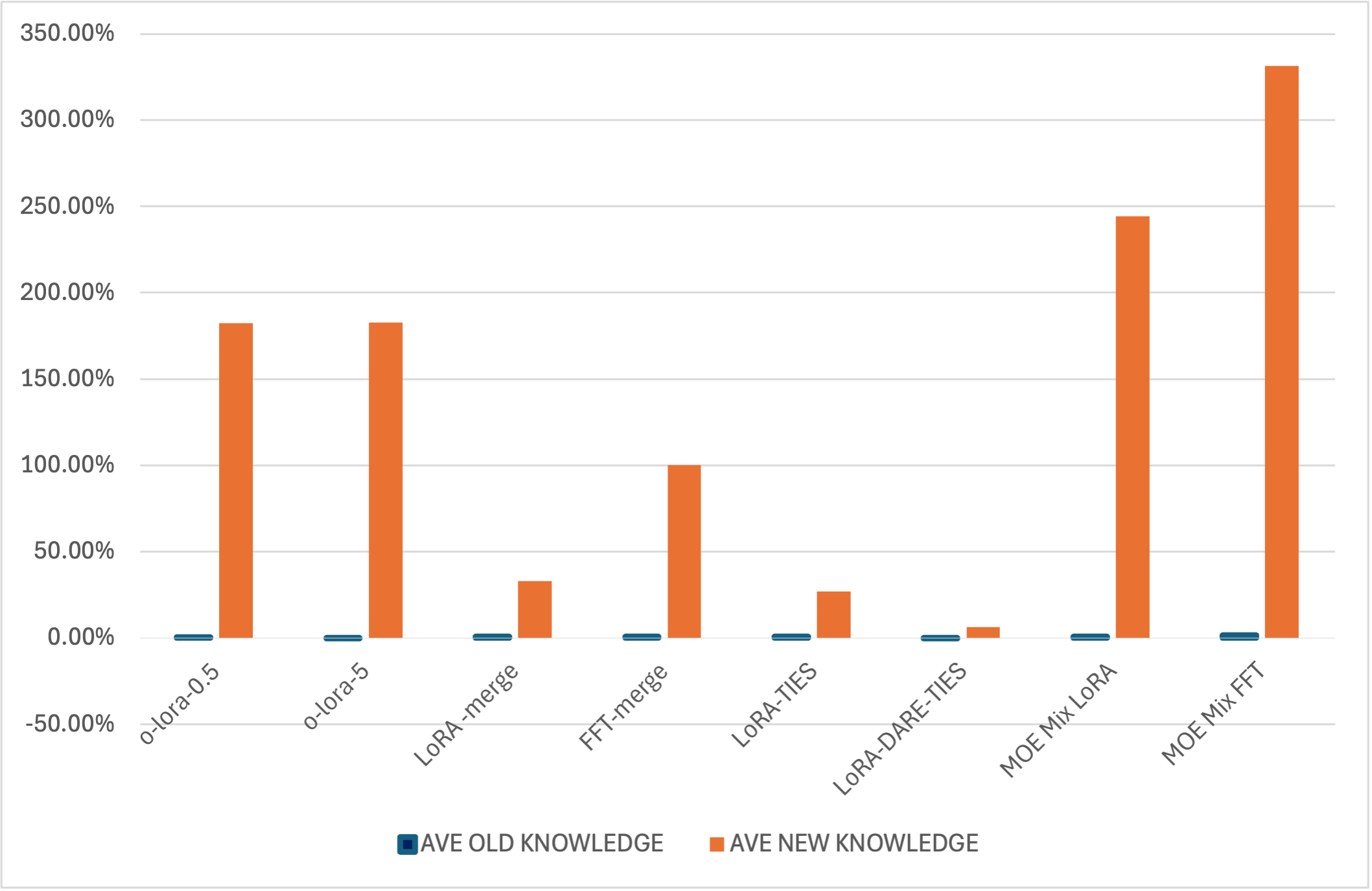}
\caption{Semi-cooperative setting. Summary comparison of the final gain from base between orthogonal subspace o-lora successive training versus model merging and MOE model mixing of the individually-trained models, both LoRA MOE and FFT MOE. MOE model mixing overall provides the most benefit by leveraging on synergies across the models. Old Knowledge sees a small improvement using most of the methods, with a small decline only using o-lora-5 and LoRA-DARE-TIES. }
\label{fig:MOEmerge-summary}
\end{figure}

A  summary is provided in Figure \ref{fig:MOEmerge-summary}. The TIES and TIES-DARE heuristics do not perform well on  new knowledge retention as compared to the other methods. We see that overall, MOE model mixing performs best, both FFT and LoRA MOE Mixing, with the orthogonal subspace o-lora variants performing well also. It should be noted that the FFT MOE Mixed model does not include all 7 of the expert modules, due to the high memory requirement of loading the significantly larger model. Rather we use 4 of the 7 experts in the FFT MOE Mixed model, namely meeting, py150, science and math, hence the average number is not fully performance on new knowledge of the MOE Mix FFT is not fully comparable to the other new knowledge averages, and care should be taken in assessing MOE Mix FFT with respect to the other methods. 

 Old Knowledge change is small but positive, reflecting improvement in Old Knowledge performance, when using all of the methods except o-lora with a large weight of 5 and LoRA-DARE-TIES. For those two methods there is a small decrease in Old Knowledge performance. 

Detailed figures are provided in the appendix in Figure \ref{fig:MOEmerge-new} for new knowledge acquisition and retention and Figure \ref{fig:MOEmerge-old} for old knowledge retention. Full fine-tuning merging via straight averaging performs quite well on old knowledge except for a sharp decline in mmlu which impacts its overall performance. 

In summary, in the semi-cooperative case, MOE model mixing and model merging of one-round models remain a very competitive alternative to sequential training. If sequential training is preferred, judicious use of orthogonal subspace o-lora stands out as a good approach.

\begin{table}[!ht]
    \centering
    \begin{tabular}{|l|l|lllllll|}
    \hline
        & \textbf{AVE} & \textbf{mmlu} & \textbf{arc\_c} & \textbf{comqa} & \textbf{hella} & \textbf{opnbk} & \textbf{piqa} & \textbf{wino} \\ \hline
        \textbf{base} & 0.6611 & 0.6223 & 0.5350 & 0.6896 & 0.7920 & 0.4500 & 0.8063 & 0.7324 \\ 
        \textbf{FFT 1} & 0.5106 & 0.2547 & 0.3746 & 0.2031 & 0.785 & 0.43 & 0.7709 & 0.7561 \\ 
        \textbf{FFT 1,2} & 0.4496 & 0.255 & 0.3208 & 0.1974 & 0.6292 & 0.37 & 0.6828 & 0.6922 \\ 
        \textbf{FFT 1-3} & 0.5383 & 0.2594 & 0.5009 & 0.2105 & 0.8097 & 0.448 & 0.8069 & 0.7324 \\ 
        \textbf{FFT 1-4} & 0.5387 & 0.2796 & 0.4983 & 0.2236 & 0.7945 & 0.464 & 0.7982 & 0.7127 \\ 
        \textbf{FFT 1-5} & 0.6109 & 0.512 & 0.4403 & 0.6716 & 0.7457 & 0.43 & 0.7709 & 0.7056 \\ 
        \textbf{FFT 1-6} & 0.6186 & 0.527 & 0.4667 & 0.6658 & 0.7599 & 0.44 & 0.7758 & 0.6953 \\ 
        \textbf{FFT 1-7} & 0.6195 & 0.5475 & 0.4343 & 0.6855 & 0.7499 & 0.43 & 0.778 & 0.7111 \\ \hline
        \textbf{gain base} & -7\% & -12.03\% & -18.82\% & -0.59\% & -5.31\% & -4.44\% & -3.51\% & -2.91\% \\ \hline
    \end{tabular}
     \caption{Full fine-tuning impact on old knowledge retention across successive tasks,   where the number 1-7 indicates the tasks in the following order cstance, fomc, meeting, py150, science, 20minuten, math. Notice that the average gain from the base model performance is a 7\% drop.}
      \label{tab:FFT}
\end{table}

\begin{table}[!ht]
    \centering
    \begin{tabular}{|l|l|lllllll|}
    \hline
         & \textbf{AVE} & \textbf{mmlu} & \textbf{arc\_c} & \textbf{comqa} & \textbf{hella} & \textbf{opnbk} & \textbf{piqa} & \textbf{wino} \\ \hline
        \textbf{LoRA  1} & 0.6642 & 0.6289 & 0.5444 & 0.6929 & 0.7934 & 0.448 & 0.8096 & 0.7324 \\ 
        \textbf{loRA  1,2} & 0.6690 & 0.6315 & 0.5512 & 0.7002 & 0.7955 & 0.454 & 0.8118 & 0.7388 \\ 
        \textbf{LoRA  1-3} & 0.6670 & 0.6256 & 0.558 & 0.683 & 0.804 & 0.454 & 0.8074 & 0.7372 \\ 
        \textbf{LoRA 1-4} & 0.6667 & 0.6268 & 0.5452 & 0.6839 & 0.8102 & 0.456 & 0.8058 & 0.7388 \\ 
        \textbf{LoRA 1-5} & 0.6551 & 0.6095 & 0.494 & 0.6781 & 0.8057 & 0.45 & 0.8052 & 0.7435 \\ 
        \textbf{LoRA 1-6} & 0.6560 & 0.6103 & 0.5119 & 0.6609 & 0.8069 & 0.448 & 0.8058 & 0.7482 \\ 
        \textbf{LoRA 1-7} & 0.6541 & 0.5974 & 0.5154 & 0.6921 & 0.781 & 0.448 & 0.8101 & 0.7348 \\ \hline
        \textbf{gain  base} & -1\% & -4.01\% & -3.66\% & 0.36\% & -1.39\% & -0.44\% & 0.47\% & 0.32\% \\ \hline
    \end{tabular}
    \caption{LoRA training impact on old knowledge retention across successive tasks, where the number 1-7 indicates the tasks in the following order cstance, fomc, meeting, py150, science, 20minuten, math. Notice that the average gain from the base model performance is a 1\% drop, smaller than the drop with FFT. }
    \label{tab:LoRA}
\end{table}

\begin{table}[!ht]
    \centering
    \begin{tabular}{|l|llllllll|l|}
    \hline
         \textbf{} &    \textbf{cstan} & \textbf{fomc} & \textbf{meet} & \textbf{py150} & \textbf{sci} & \textbf{20min} & \textbf{gsm8k} & \textbf{gsmcot} &  \textbf{AVE }\\ \hline
        \textbf{FFT 1 } & 0.511 & 0.4839 & 0.0444 & 0.3033 & 0.115 & 0.3709 & 0.0008 & 0.0713 & 0.2376 \\
        \textbf{FFT 1,2 } & 0.444 & 0.7157 & 0.0073 & 0.0032 & 0.1265 & 0.3701 & 0 & 0 & 0.2084 \\ 
        \textbf{FFT 1-3 } & 0.457 & 0.6109 & 0.39 & 0.2256 & 0.217 & 0.3608 & 0.1812 & 0.2714 & 0.3392 \\ 
        \textbf{FFT 1-4 } & 0.459 & 0.619 & 0.3709 & 0.5238 & 0.2185 & 0.3651 & 0.1099 & 0.232 & 0.3623 \\ 
        \textbf{FFT 1-5 } & 0.448 & 0.4657 & 0.4013 & 0.5248 & 0.922 & 0.3512 & 0.0857 & 0.1296 & 0.4160 \\ 
        \textbf{FFT 1-6 } & 0.461 & 0.4657 & 0.3927 & 0.5211 & 0.9255 & 0.4078 & 0.1266 & 0.1357 & 0.4295 \\ 
        \textbf{FFT 1-7 } & 0.457 & 0.4133 & 0.3651 & 0.4924 & 0.917 & 0.3896 & 0.7293 & 0.7384 & 0.5628 \\ \hline
        \textbf{gain  base } & 110\% & 61\% & 119\% & 985\% & 26\% & 11\% & 44\% & 34\% & 60\% \\ \hline
        \textbf{BWT} & -11\% & -42\% & -6\% & -6\% & -1\% & -4\% & NA & NA & -- \\ \hline
    \end{tabular}
     \caption{FFT training impact on new knowledge acquisition and retention across successive tasks, where the number 1-7 indicates the tasks in the following order cstance, fomc, meeting, py150, science, 20minuten, math. }
     \label{tab:newFFT}
\end{table}

\begin{table}[!ht]
    \centering
    \begin{tabular}{|l|llllllll|l|}
    \hline
         \textbf{} &    \textbf{cstan} & \textbf{fomc} & \textbf{meet} & \textbf{py150} & \textbf{sci} & \textbf{20min} & \textbf{gsm8k} & \textbf{gsmcot} &  \textbf{AVE }\\ \hline
        \textbf{LoRA  1 } & 0.535 & 0.5081 & 0.1783 & 0.0585 & 0.7545 & 0.3556 & 0.5133 & 0.5656 & 0.434 \\ 
        \textbf{LoRA  1,2 } & 0.5175 & 0.7036 & 0.1745 & 0.063 & 0.7615 & 0.3624 & 0.5171 & 0.5618 & 0.458 \\ 
        \textbf{LoRA  1-3 } & 0.532 & 0.6915 & 0.3771 & 0.067 & 0.749 & 0.3515 & 0.5034 & 0.5444 & 0.477 \\ 
        \textbf{LoRA 1-4 } & 0.515 & 0.6815 & 0.3463 & 0.5438 & 0.7065 & 0.3641 & 0.4882 & 0.5625 & 0.526 \\ 
        \textbf{LoRA 1-5 } & 0.497 & 0.6613 & 0.3415 & 0.5324 & 0.8975 & 0.3538 & 0.3434 & 0.4018 & 0.504 \\ 
        \textbf{LoRA 1-6 } & 0.5205 & 0.6331 & 0.3329 & 0.5351 & 0.8895 & 0.4068 & 0.4117 & 0.3973 & 0.516 \\ 
        \textbf{LoRA 1-7 } & 0.508 & 0.5988 & 0.348 & 0.5437 & 0.531 & 0.3903 & 0.7096 & 0.7104 & 0.543 \\ \hline
        \textbf{gain  base } & 133\% & 134\% & 109\% & 1098\% & -27\% & 11\% & 41\% & 29\% & 54\% \\ \hline
        \textbf{BWT} & -5\% & -15\% & -8\% & 0\% & -41\% & -4\% & NA & NA & -- \\ \hline
    \end{tabular}
      \caption{LoRA training impact on new knowledge acquisition and retention across successive tasks, where the number 1-7 indicates the tasks in the following order cstance, fomc, meeting, py150, science, 20minuten, math. }
     \label{tab:newLoRA}
\end{table}

%%%%%%%%%%%%%%%%

\subsection{Fully-Cooperative  Model Training}

In the fully-cooperative setting, one has access to the datasets used in training the already-trained models. Thus, joint training over all datasets, and sequential training with replay, in which a small fraction of earlier datasets is added to each successive training, become feasible training paradigms. We examine both approaches here.

Joint training is a straightforward way to make use of increasingly larger training data, and can be seen as mirroring the approach used in pre-training models: each generation of new models is pre-trained on increasingly larger datasets, and indeed provide better performance. One may expect the same to hold with joint fine-tuning over all data sets. 

Replay is a more economical approach that remains sequential in nature. As before, the training on a given task starts with the model trained on the previous task. We use the same task ordering as before. However, 10\% of the training data set size is taken randomly from each of the previous tasks. Note that for the first 6 tasks, this represents $0.1*5000=500$ examples from each of the TRACE datasets seen already. On the other hand, for the 7th task, as the math dataset has 395K examples, the TRACE datasets are upsampled to provide each $0.1*395K=39.5K$ examples to the math training. Similarly, for joint training, the smaller TRACE datasets are upsampled so that the probability of selection across all datasets is uniform.

\begin{figure}[ht]
\centering
\includegraphics[width=0.85\linewidth]{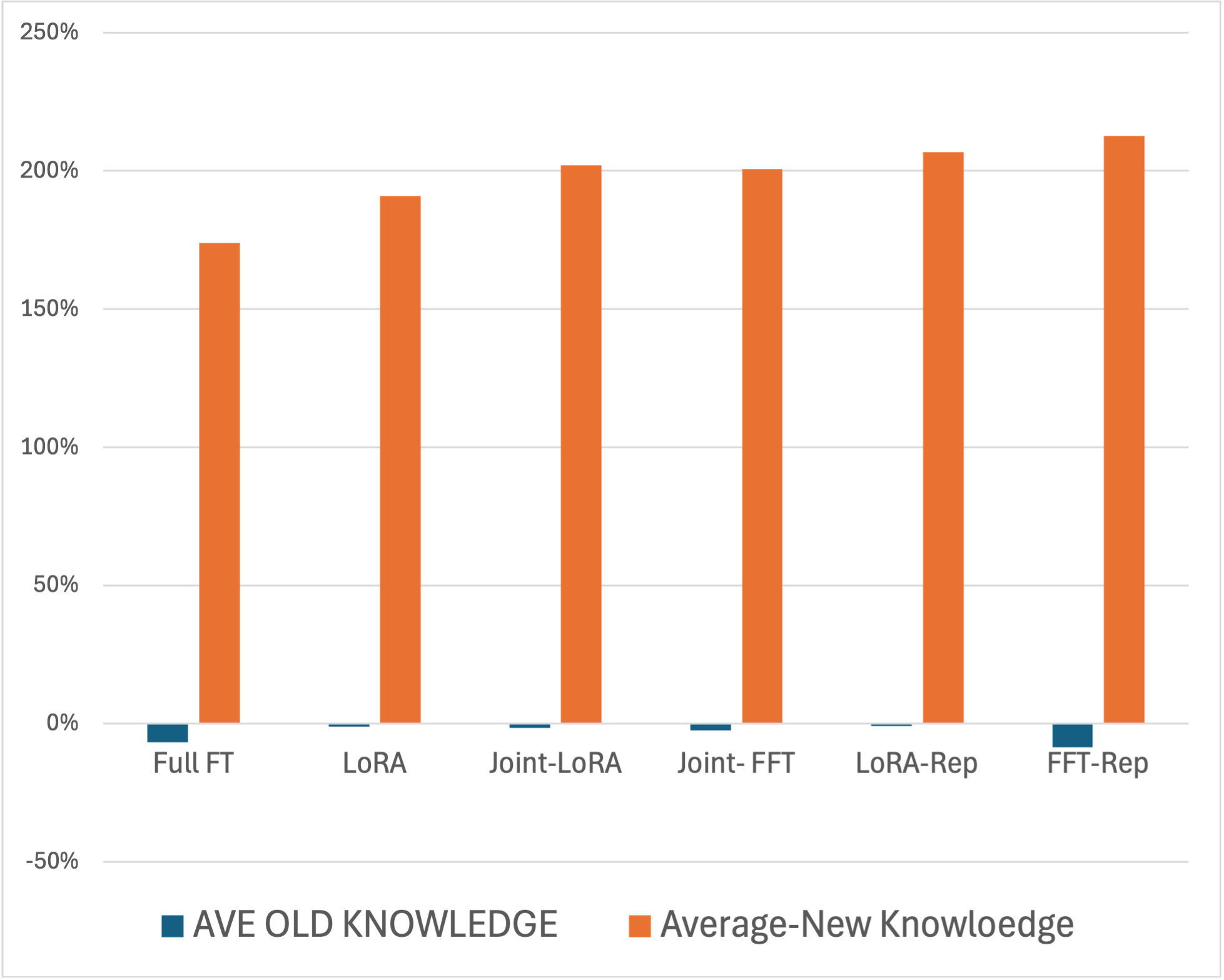}
\caption{Fully-cooperative setting, comparison between sequential FFT, LoRA, and the same with 10\% replay, and joint FFT and LoRA training. Sequential LoRA training with replay perform best on new knowledge acquisition and old knowledge retention. }
\label{fig:fully}
\end{figure}

The results are summarized in Figure \ref{fig:fully}. While one might expect joint training to outperform all other methods, sequential training with replay actually performs better than joint training. LoRA performs better overall in retaining old knowledge in this fully-cooperative setting. The negative blue-color bars are nearly imperceptible, meaning very little loss of old knowledge using LoRA.
The details are presented in the Appendix in Table \ref{tab:fully-old} for old knowledge retention and Table \ref{tab:fully-new} for new knowledge acquisition and retention. Amongst the LoRA techniques in the fully-cooperative setting, sequential LoRA with replay offers the best performance in retaining old knowledge.

In terms of new knowledge acquisition (and retention, for sequential learning), sequential LoRA with replay and FFT with replay both perform very well, with joint training slightly inferior. A look at the detailed results, however, in Table \ref{tab:fully-new} in the Appendix, shows that FFT with replay strongly outperforms LoRA with replay on the math  and science tests.  

Overall, however, in the fully-cooperative setting joint training represents a good and reliable method, and joint LoRA training has the advantage of less old knowledge degradation. Nonetheless, sequential training with replay is an adequate alternative and can be more economical in terms of growing datasize.

%%%%%%%%%%%%%%%%%
\section{Conclusion}
\label{sec:conc}
We analyze collaboratively adding new knowledge to a large language model. 
In the semi-cooperative setting where datasets are not available after training, MOE mixing, model merging, and LoRA-based orthogonal subspace sequential learning perform well. It is preferable to use a small weight on the orthogonality term in the loss function to mitigate the reduction  in CE loss optimization that comes from the presence of that term. 

LoRA overall performs better in most cases than full-fine tuning of all parameters when both new knowledge acquisition and old, including recent, knowledge retention are taken into account. In the fully-cooperative setting, joint training and sequential training with replay are both effective approaches with LoRA training  generally preferable to full fine-tuning.

The conclusions of this work are heartening in that LoRA has become a de-facto approach for model fine-tuning due to its considerable computational advantages combined with the plug-and-play nature of the resulting adapters. This work shows that LoRA is not only an efficient and practical solution to fine-tuning but also an effective one for collaborative and sequential training.

The high quality of MOE mixing is also notable. The method, like LoRA training, is distinguished by its computational efficiency. Given trained models the MOE can be created with little to no training using the approach provided in \cite{moe}. Yet, it is not only efficient but highly effective in leveraging multiple trained models or adapters.

\section*{Acknowledgements} The authors would like to acknowledge Shiau Hong Lim for his deep insights and  discussions throughout this work and Ronny Luss for his helpful comments on the paper.

%\newpage
\bibliographystyle{naturemag}
\bibliography{refs}
%\newpage

%\medskip
%\printbibliography

\section{Appendix}

\begin{figure}[ht]
\centering
\includegraphics[width=0.85\linewidth]{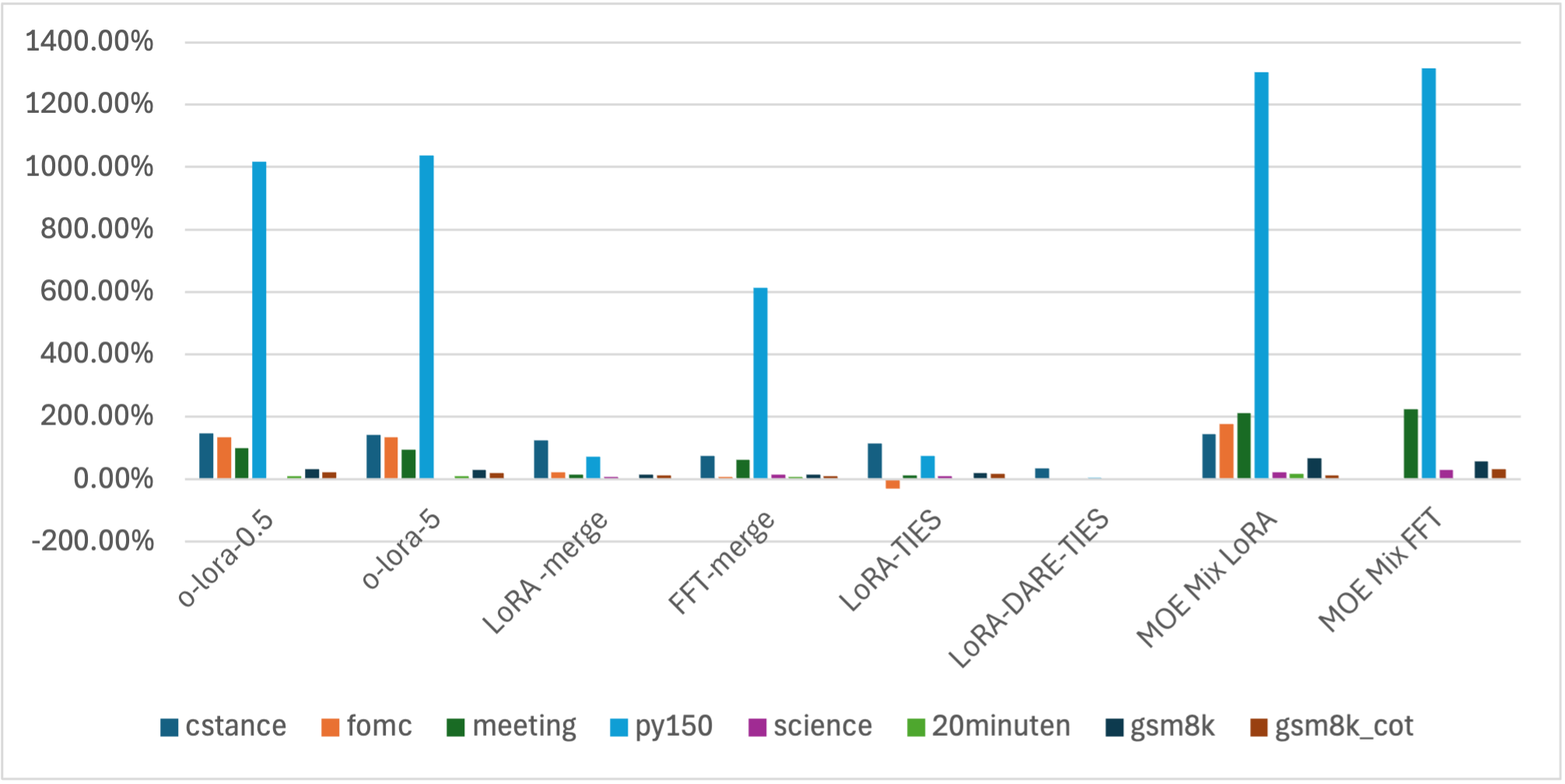}
\caption{Comparison of the final gain from base between orthogonal subspace o-lora successive training versus model merging and MOE model mixing of the individually-trained models on new knowledge. While the results are dominated by performance on the python coding task, we do see that MOE model mixing overall provides the most benefit with o-lora in second place, as shown also in the summary in Figure \ref{fig:MOEmerge-summary}.}
\label{fig:MOEmerge-new}
\end{figure}

\begin{figure}[ht]
\centering
\includegraphics[width=0.85\linewidth]{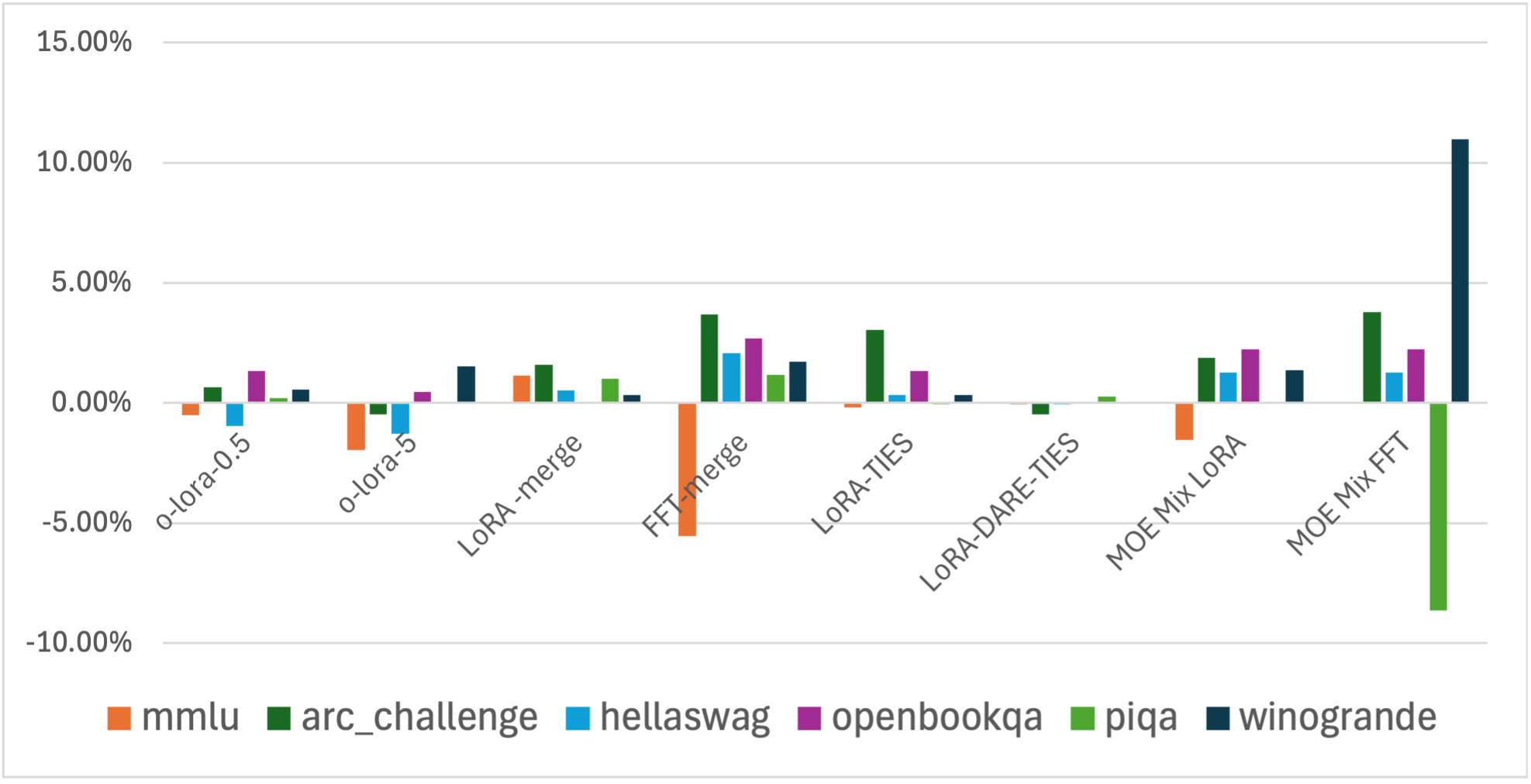}
\caption{Comparison of the final gain from base between orthogonal subspace o-lora successive training versus model merging and MOE model mixing of the individually-trained models on old knowledge retention. Here, the results are dominated by the degradation in the mmlu task, which penalises FFT merging in particular; without that poor result, FFT merging  performs quite well on old knowledge retention.  MOE model mixing also loses performance on mmlu, but even a detailed view shows that overall MOE mixing provides the most benefit, as shown also in the summary in Figure \ref{fig:MOEmerge-summary}.}
\label{fig:MOEmerge-old}
\end{figure}

We provide additional results here. In Figure \ref{fig:MOEmerge-new} we examine the final gain from the base model after 7 rounds of training in the semi-cooperative setting in terms of new and recently-acquired knowledge. The analogous results on old knowledge are provided in Figure \ref{fig:MOEmerge-old}.

Table \ref{tab:fully-old} provides the detailed results from the fully-cooperative setting on old knowledge while Table \ref{tab:fully-new} provides the analogous detailed results on new and recent knowledge. Recall that due to the larger size of the MOE Mix FFT model, only 4 expert modules are included, and as such comparison between MOE Mix FFT and all of the other methods should be made with caution.

\begin{table}[!ht]
    \centering
    \begin{tabular}{|l|l|l|l|l|l|l|l|l|}
    \hline   
    & \textbf{AVE} & \textbf{mmlu} & \textbf{arc\_c} & \textbf{comqa} & \textbf{hella} & \textbf{opnbk} & \textbf{piqa} & \textbf{wino} 
 \\ \hline
        \textbf{base } & 0.6611 & 0.6223 & 0.5350 & 0.6896 & 0.7920 & 0.4500 & 0.8063 & 0.7324 \\ \hline
        \textbf{Joint-LoRA } & 0.6524 & 0.6154 & 0.4633 & 0.7043 & 0.7933 & 0.454 & 0.7933 & 0.7435 \\ \hline
        \textbf{LoRA } & 0.6541 & 0.5974 & 0.5154 & 0.6921 & 0.781 & 0.448 & 0.8101 & 0.7348 \\ \hline
        \textbf{LoRA-Rep } & 0.6556 & 0.6254 & 0.4966 & 0.6806 & 0.7967 & 0.464 & 0.7987 & 0.7269 \\ \hline
        \textbf{Joint-FFT } & 0.6429 & 0.5706 & 0.5111 & 0.6757 & 0.7521 & 0.468 & 0.7835 & 0.7395 \\ \hline
        \textbf{FFT } & 0.6195 & 0.5475 & 0.4343 & 0.6855 & 0.7499 & 0.43 & 0.778 & 0.7111 \\ \hline
        \textbf{FFT-Rep} & 0.6067 & 0.5243 & 0.4121 & 0.6765 & 0.748 & 0.444 & 0.7622 & 0.6796 \\ \hline
    \end{tabular}
    \caption{Details of the results in the fully-cooperative setting, evaluated on old knowledge retention.}
    \label{tab:fully-old}
\end{table}

\begin{table}[!ht]
    \centering
    \begin{tabular}{|l|l|l|l|l|l|l|l|l|l|l|}
    \hline
 \textbf{} &    \textbf{cstan} & \textbf{fomc} & \textbf{meet} & \textbf{py150} & \textbf{sci} & \textbf{20min} & \textbf{gsm8k} & \textbf{gsmcot} &  \textbf{AVnew }&  \textbf{AVall }
 \\ \hline
        \textbf{base } & 0.218 & 0.256 & 0.167 & 0.045 & 0.726 & 0.351 & 0.505 & 0.550 & 0.352 & 0.496 \\ \hline
        \textbf{Jnt-LoR } & 0.519 & 0.716 & 0.375 & 0.545 & 0.894 & 0.408 & 0.608 & 0.616 & 0.585 & 0.617 \\ \hline
        \textbf{LoRA } & 0.508 & 0.599 & 0.348 & 0.544 & 0.531 & 0.390 & 0.710 & 0.710 & 0.595 & 0.543 \\ \hline
        \textbf{LoR-Rep } & 0.552 & 0.724 & 0.369 & 0.540 & 0.909 & 0.409 & 0.716 & 0.684 & 0.613 & 0.633 \\ \hline
        \textbf{Jnt-FFT } & 0.508 & 0.692 & 0.398 & 0.527 & 0.941 & 0.411 & 0.671 & 0.673 & 0.603 & 0.621 \\ \hline
        \textbf{FFT } & 0.457 & 0.413 & 0.365 & 0.492 & 0.917 & 0.390 & 0.729 & 0.738 & 0.589 & 0.563 \\ \hline
        \textbf{FFT-Rep} & 0.528 & 0.704 & 0.465 & 0.535 & 0.941 & 0.413 & 0.726 & 0.754 & 0.621 & 0.633 \\ \hline
    \end{tabular}
    \caption{Details of the results in the fully-cooperative setting, evaluated on new knowledge acquisition and retention.}
    \label{tab:fully-new}
\end{table}

\end{document}